\documentclass[english]{scrartcl}
\usepackage[T1]{fontenc}
\usepackage[latin9]{inputenc}
\usepackage{geometry}
\usepackage{color}
\usepackage{babel}
\usepackage{array}
\usepackage{verbatim}
\usepackage{amsmath}
\usepackage{amsthm}
\usepackage[authoryear]{natbib}
\usepackage[unicode=true,pdfusetitle,
 bookmarks=true,bookmarksnumbered=false,bookmarksopen=false,
 breaklinks=false,pdfborder={0 0 1},backref=false,colorlinks=true]
 {hyperref}
\hypersetup{
 citecolor=blue}

\makeatletter

\providecommand{\tabularnewline}{\\}

\newcommand{\lyxaddress}[1]{
	\par {\raggedright #1
	\vspace{1.4em}
	\noindent\par}
}
\theoremstyle{definition}
\newtheorem{defn}{\protect\definitionname}
\theoremstyle{plain}
\newtheorem{thm}{\protect\theoremname}

\usepackage{pgfplots}
\usepackage{amsmath}
\usetikzlibrary{calc}
\usetikzlibrary{shapes.geometric}
\usetikzlibrary{decorations.markings}
\usetikzlibrary{positioning}
\usepackage{caption}
\makeatother

\providecommand{\definitionname}{Definition}
\providecommand{\theoremname}{Theorem}

\begin{document}
\title{Algorithmic syntactic causal identification}
\author{Dhurim Cakiqi$^{\star}$, Max A. Little$^{\star}$}
\maketitle

\lyxaddress{\begin{center}
$^{\star}$School of Computer Science, University of Birmingham,
UK
\par\end{center}}
\begin{abstract}
Causal identification in causal Bayes nets (CBNs) is an important
tool in causal inference allowing the derivation of interventional
distributions from observational distributions where this is possible
in principle. However, most existing formulations of causal identification
using techniques such as d-separation and do-calculus are expressed
within the mathematical language of classical probability theory on
CBNs. However, there are many causal settings where probability theory
and hence current causal identification techniques are inapplicable
such as relational databases, dataflow programs such as hardware description
languages, distributed systems and most modern machine learning algorithms.
We show that this restriction can be lifted by replacing the use of
classical probability theory with the alternative axiomatic foundation
of symmetric monoidal categories. In this alternative axiomatization,
we show how an unambiguous and clean distinction can be drawn between
the general \emph{syntax }of causal models and any specific \emph{semantic
}implementation of that causal model. This allows a \emph{purely syntactic
}algorithmic description of general causal identification by a translation
of recent formulations of the general ID algorithm through \emph{fixing}.
Our description is given entirely in terms of the non-parametric ADMG
structure specifying a causal model and the algebraic signature of
the corresponding monoidal category, to which a sequence of manipulations
is then applied so as to arrive at a modified monoidal category in
which the desired, purely syntactic interventional causal model, is
obtained. We use this idea to derive purely syntactic analogues of
classical back-door and front-door causal adjustment, and illustrate
an application to a more complex causal model.
\end{abstract}

\section{Introduction}

\emph{Causal Bayes nets }(CBNs)\emph{ }are probabilistic models in
which \emph{causal influences }between \emph{random variables} are
expressed via the use of \emph{graphs} with \emph{nodes }in these
graphs being the random variables and \emph{directed edges }indicating
the direction of causality between them \citep{Pearl:2009,Bareinboim:2020}.
Every such\emph{ directed acyclic graph }(DAG) with \emph{latent }(\emph{unobserved})
nodes has a corresponding\emph{ acyclic directed mixed graph }(ADMG)
which is obtained from the DAG through \emph{latent projection }which
simplifies the DAG whilst preserving its \emph{causal} \emph{d-separation
}properties \citep{Richardson:2012,Pearl:2009}.

The ADMG $\mathcal{G}=\left(\mathbf{V}^{\mathcal{G}},\mathbf{E}^{\mathcal{G}}\right)$
with variables $\mathbf{V}^{\mathcal{G}}$ and edges $\mathbf{E}^{\mathcal{G}}$
has bidirected edges indicating \emph{unmeasured confounding} \citep{Richardson:2012}.
Excluding the bidirected edges, the topological properties of these
graphs are given by their set-valued \emph{parent} $\mathrm{pa}_{\mathcal{G}}$
and \emph{child }functions $\mathrm{ch}_{\mathcal{G}}$. \emph{Ancestors
}$\mathrm{an}_{\mathcal{G}}$ and \emph{descendents }$\mathrm{de}_{\mathcal{G}}$
are determined recursively from these. The \emph{subgraph} $\mathcal{G}_{\mathbf{Y}}$
for $\mathbf{Y}\subset\mathbf{V}^{\mathcal{G}}$ is obtained by deleting
from $\mathbf{V}^{\mathcal{G}}$ all nodes not in $\mathbf{Y}$ and
edges which connect to those removed variables. A sequence of the
subset of the nodes in $\mathbf{V}^{\mathcal{G}}$ such that every
node in that sequence occurs before its children (or after its parents)
in $\mathcal{G}$ through unidirectional edges, is called a \emph{topological
ordering }of the ADMG \citep{Bareinboim:2020}. For such an ordering
the function $\mathrm{pre}_{\mathcal{G}}\left(V\right)$ gives the
set of all the nodes before $V$ in the sequence, and $\mathrm{succ}_{\mathcal{G}}\left(V\right)$
gives all the nodes after it in the sequence. In the ADMG, sets of
variables which are connected through a sequence of bidirected edges
are called \emph{districts} $\mathrm{dis}_{\mathcal{G}}$ \citep{Richardson:2012}.
Nodes $V\in\mathbf{V}^{\mathcal{G}}$ which have the property that
$\mathrm{dis}_{\mathcal{G}}\left(V\right)\cap\mathrm{de}_{\mathcal{G}}\left(V\right)=\left\{ V\right\} $
are called \emph{fixable }nodes \citep{Richardson:2012}.

For an ADMG with no bidirected edges (thus, no latent variables, equivalent
to a CBN over a DAG), it is always possible to derive any interventional
distribution from the joint distribution over the variables in the
DAG using the \emph{truncated factorization }\citep{Pearl:2009}.
However, more generally, in the presence of unobserved confounding
(e.g. models having bidirected edges in the ADMG) this is no longer
true and only certain interventional distributions can be derived
from the observed variables \citep{Shpitser:2008,Bareinboim:2020}.
Pearl's \emph{do-calculus }\citep{Pearl:2009} is a set of three algebraic
distribution transformations which it has been shown are necessary
and sufficient for deriving the interventional distribution where
this is possible \citep{Shpitser:2008}. These algebraic transformations
can be applied ad-hoc or, more systematically, using Shpitser's \emph{ID
algorithm} to derive a desired interventional distribution \citep{Shpitser:2008}.
More recently, the specific conditions under which any particular
interventional distribution can be determined from the observed variables
using do-calculus or some other systematic algorithm, has been simplified
in terms of \emph{fixing operations }and \emph{reachable subgraphs}
in causal ADMGs \citep{Richardson:2012}. Exploiting the same reasoning,
\citet{Richardson:2012} show how fixing operations can be combined
in a simple algorithm which achieves the same result.

This algorithm, as with most algorithms for causal inference, is expressed
in terms of CBNs using random variables and classical probabilities
where probabilistic conditioning indicates the direction of causal
inference in an ADMG. Such causal identification algorithms rely on
simultaneous manipulation of the ADMG, tracking the consequence of
such manipulations on the corresponding (joint) distribution over
that graph. As long as the appropriate \emph{Markov property} holds
\citep{Bareinboim:2020}, which guarantees the consistency of the
distribution with the CBN, then this is a valid procedure for deriving
the desired interventional distribution. Nonetheless, there are many
practical settings where probabilistic modelling is inappropriate,
such as relational databases \citep{patterson2017knowledge}, hardware
description languages, distributed systems modelled by Petri nets
and most modern machine learning algorithms \citep{little2019machine}.
In these settings there is no such Markov property therefore it appears
that the existing causal identification algorithms are inapplicable
in these wider, non-probabilistic applications.

A different and more recently explored direction which might circumvent
this limitation is to change the fundamental axiomatic basis of the
modelling language to use (\emph{monoidal})\emph{ category theory
}instead. This amounts to a fundamental reformulation of CBNs that,
rather than organizing causal models around sets, measure theory and
graph topology which requires the additional complexity of Markov
properties to bind these together, instead views CBNs from the simpler
and more abstract vantage point of \emph{structured compositional
processes}. Causal modelling and inference in terms of \emph{string
diagrams }representing such processes has shown considerable promise.
Building on work by \citet{fong2013causal}, \citet{Cho:2019} formulated
the essential concepts of Bayesian reasoning as strings, following
which \citet{Jacobs:2021} provided an exposition of causal identification
under a slightly extended form of the \emph{front-door }causal scenario
for \emph{affine Markov categories }\citep{Fritz:2020}. Since then,
string analogues of do-calculus and d-separation have been described
\citep{Yin:2022,JMLR:v24:22-0916} and explicit description of extensions
of the categorical string diagram approach to causal modelling in
non-probabilistic settings such as machine learning \citep{Cakiqi:2022}.

Nonetheless the full promise of this reformulation has yet to be realized.
For instance, causal inference in string diagrams has, to date, only
been described in probabilistic categories for single variable interventions
in discrete sample spaces where interventions can be modelled by discrete
uniform distributions \citep{Jacobs:2021}, or more generally to string
diagrams where causal identification beyond the slightly extended
form of the front-door causal scenario is not carried out systematically
(algorithmically) but instead requires manual string manipulation
\citep{Lorenz:2023}. Thus, our goal in this report is to provide
the first, purely syntactic, algorithmic characterization of general
causal identification by fixing which is applicable to the full range
of causal models expressible as a structured, categorical compositional
process.

\section{Theory}

\subsection{Symmetric monoidal categories and their algebraic signatures}

\emph{Symmetric monoidal categories} (SMCs) are algebraic structures
which capture the notion of simultaneous \emph{sequential} and (in
our application) \emph{parallel} \emph{composition} of maps between
types. Examples of such categories include ordinary sets and functions
between these sets with the cartesian product indicating parallel
composition, the category of sets and relations \citep{fong2013causal},\emph{
Markov categories} of sample spaces with conditional distributions
modelled by sets and \emph{probability monads }between them \citep{Fritz:2020}
or other non-deterministic monads in arbitrary semifields \citep{Cakiqi:2022}.
Following \citet{Sellinger:2011}, a symmetric monoidal category \emph{signature}
$\Sigma$ provides all the information required to specify a particular
SMC. For those unfamiliar with category theory, \citet{riehl2017category}
is an excellent introduction. 
\begin{defn}
(Symmetric monoidal signature). A symmetric monoidal signature $\Sigma$,
consists of a set of object terms $\Sigma_{0}$ and morphism variables,
$\Sigma_{1}$ along with a pair of functions $\text{dom},\text{cod}:\Sigma_{1}\to\text{Mon}\left(\Sigma_{0}\right)$
which determine the domain and codomain of the morphism variables
respectively. Here, $\text{Mon}\left(\Sigma_{0}\right)=\left(\Sigma_{0},\otimes,1\right)$
is the free commutative monoid generated by the object terms in $\Sigma_{0}$. 
\end{defn}
For brevity we will use exponential notation to indicate terms in
$\text{\ensuremath{\mathrm{Mon}\left(\Sigma_{0}\right)}}$ with repeated
objects, i.e. for $A,B\in\Sigma_{0}$ the object expression $\left(A\otimes B\right)\otimes\left(1\otimes A\right)$
is written $A^{2}B$. A morphism with no input, $v:1\to A$, has monoidal
unit domain $1$; a morphism with deleted (empty) output has type
$u:A\to1$. We include the domain and codomain functions as a part
of the signature i.e. $\Sigma=\left(\Sigma_{0},\Sigma_{1},\mathrm{dom},\mathrm{cod}\right)$.
To save space, we will alternatively denote the information in $\mathrm{dom},\mathrm{cod}$
through the more traditional type specifications of the morphisms,
e.g. for the category containing the morphism $\mathrm{dom}\left(u\right)=A$,
$\mathrm{dom}\left(v\right)=AB$, $\mathrm{cod}\left(u\right)=1$
and $\mathrm{cod}\left(v\right)=V$ we write $\Sigma=\left(\left\{ A,B\right\} ,\left\{ u,v\right\} ,\left\{ u:A\to1,v:AB\to V\right\} \right)$.

An (affine, symmetric) monoidal signature $\Sigma$ determines a symmetric
monoidal category whose morphisms are \emph{generated} by combining
morphism variables using sequential composition $\cdot$ and commutative
monoidal product $\otimes$ along with identities $\mathit{id}_{A}:A\to A$,
copies $\Delta_{A}:A\to A^{2}$ and deletions $\mathit{del}_{A}:A\to1$
for every object in $A\in\Sigma_{0}$. Every expression formed this
way is itself a morphism in the category specified by $\Sigma$.

\begin{figure}
\begin{centering}
\begin{tabular}{ll}
\begin{tikzpicture}[unit length/.code={{\newdimen\tikzunit}\setlength{\tikzunit}{#1}},unit length=8mm,x=\tikzunit,y=\tikzunit,semithick,box/.style={rectangle,draw,solid,rounded corners},junction/.style={circle,draw,fill,inner sep=0},outer box/.style={draw=none},wire/.style={draw,postaction={decorate},decoration={markings,mark=at position 0.5 with {#1}}}]
\node[box,minimum size=1\tikzunit] (x) at (0.5,2) {$x_{1}$};
\node (a) at (1,0) {(a)};
\node[box,minimum height=1\tikzunit,minimum width=2\tikzunit] (x3) at (1,6) {$x_{3}$};
\node[box,minimum size=1\tikzunit] (x4) at (0.5,4) {$x_{2}$};
\node[junction,minimum size=0.25\tikzunit] (xcopy) at (0.5,3) {};
\node (Y) at (1,7.5) {$X_{3}$};
\node (X) at (2,5) {$X_{1}$};
\node (Z) at (0,5) {$X_{2}$};
\path[wire] (x.north) to[out=90, in=-90] (xcopy.south);
\path[wire] (xcopy.north) to[out=90, in=-90]  (x4.south);
\path[wire] (x4.north)  to[out=90,in=-90] (0.5,5.5) to[out=90, in=-90] ($(x3.south)-(0.5,0)$);
\path[wire] (xcopy.east)  to[out=0, in=-90] (1.5,4) to[out=90, in=-90] ($(x3.south)+(0.5,0)$);
\path[wire] (x3.north) to (Y);
\end{tikzpicture} & \begin{tikzpicture}[unit length/.code={{\newdimen\tikzunit}\setlength{\tikzunit}{#1}},unit length=8mm,x=\tikzunit,y=\tikzunit,semithick,box/.style={rectangle,draw,solid,rounded corners},junction/.style={circle,draw,fill,inner sep=0},outer box/.style={draw=none},wire/.style={draw,postaction={decorate},decoration={markings,mark=at position 0.5 with {#1}}}]
\node[box,minimum height=1\tikzunit,minimum width=1\tikzunit] (x3) at (2,1.3) {$q$};
\node (Y) at (2,4.7) {$X_{3}$};
\node (a) at (2,-2.75) {(b)}; 
\path[wire] (x3.north) to[out=90, in=-90] (Y) ;
\end{tikzpicture}\tabularnewline
\end{tabular}
\par\end{centering}
\caption{Example string diagram representations of maximal causal models for
monoidal signatures: (a) signature $\Sigma=\left(\left\{ X_{1},X_{2},X_{3}\right\} ,\left\{ x_{1},x_{2},x_{3}\right\} ,\left\{ x_{1}:1\to X_{1}^{2},x_{2}:X_{1}\to X_{2},x_{3}:X_{1}X_{2}\to X_{3}\right\} \right)$
with explicit internal causal mechanisms; (b) exterior signature $\mathrm{Ext}\left(\Sigma\right)=\left(\left\{ X_{3}\right\} ,\left\{ q\right\} ,\left\{ q:1\to X_{3}\right\} \right)$
hiding the internal causal mechanisms in (a).\label{fig:example-strings}}
\end{figure}
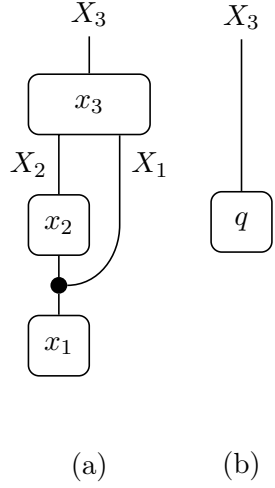
However, the signature also determines a \emph{specific }causal model
which is the one of practical interest, in the following sense. Construct
an expression in which all causal module morphism variable terms appear
once, composed using sequential composition $\circ$ and monoidal
product $\otimes$, inserting only the necessary identities and copies
in order to ensure that the domains and codomains of these morphisms
are matched. We call this expression the \emph{maximal model }(quotiented
by the identities, copies and deletions) determined by the signature.
The maximal model is a (possibly composite) causal morphism in the
monoidal category, with its own domain and codomain. For example,
the signature
\begin{equation}
\Sigma=\left(\left\{ X_{1},X_{2},X_{3}\right\} ,\left\{ x_{1},x_{2},x_{3}\right\} ,\left\{ x_{1}:1\to X_{1}^{2},x_{2}:X_{1}\to X_{2},x_{3}:X_{1}X_{2}\to X_{3}\right\} \right),\label{eq:signature-example}
\end{equation}
 has the maximal causal model expression $q=x_{3}\cdot\left(x_{2}\otimes\mathit{id}_{X_{1}}\right)\cdot x_{1}$
with type
\begin{equation}
\mathrm{dom}\left(q\right)=1\overset{x_{1}}{\to}X_{1}X_{1}\overset{x_{2}\otimes\mathit{id}_{X_{1}}}{\longrightarrow}X_{2}X_{1}\overset{x_{3}}{\to}X_{3}=\mathrm{cod}\left(q\right).
\end{equation}

While the expression $q$ is itself a morphism in the category generated
by $\Sigma$, below we will find it useful to isolate this as a separate
causal morphism which generates a category with composite signature
$\Sigma^{\prime}=\left(\left\{ X_{3}\right\} ,\left\{ q\right\} ,\left\{ q:1\to X_{3}\right\} \right)$.
This signature hides the \emph{internal }details of how $q$ was obtained;
the detailed signature can always be reconstructed from information
in the expression from which it is formed. We call the signature $\Sigma^{\prime}=\mathrm{Ext}\left(\Sigma\right)$
the \emph{exterior }signature $\Sigma$. Because of the modular nature
of causal ADMGs, in the corresponding monoidal signature there is
a one-one mapping between domain object labels and morphism labels.
\begin{defn}
(Causal module labels). For a given monoidal signature $\Sigma$ representing
an ADMG with objects and morphism labels $\Sigma_{0},\Sigma_{1}$
the function $\text{Module}:\Sigma_{0}\to\Sigma_{1}$, returns the
morphism label associated with a causal module. 
\end{defn}
As an example, in the case of the exterior signature above $\mathrm{Module}\left(X_{3}\right)=q$
because in $\mathrm{Ext}\left(\Sigma\right)$, $\mathrm{cod}\left(q\right)=X_{3}$.

A convenient graphical notational device for representing such (maximal)
causal models are \emph{string diagrams }\citep{Sellinger:2011} which
have a rigorous algebraic meaning which coincides with that of monoidal
categories, see Figure \ref{fig:example-strings}. We recommend \citet{Coecke_Kissinger_2017}
as a background to string diagrams for those unfamiliar with the concept. 

In the next section we build a bridge between latent CBNs represented
as ADMGs, and their representation as (affine) SMC signatures specifying
a causal model.

\subsection{Monoidal category signatures from ADMGs}

The monoidal category signature $\Sigma^{\mathcal{G}}$ generated
by the ADMG $\mathcal{G}$ is $\Sigma^{\mathcal{G}}=\left(\Sigma_{0}^{\mathcal{G}},\Sigma_{1}^{\mathcal{G}},\mathrm{dom},\mathrm{cod}\right)$,
where $\Sigma_{0}^{\mathcal{G}}=\mathbf{V}^{\mathcal{G}}$, $\Sigma_{1}^{\mathcal{G}}=\left\{ \mathrm{Module}\left(V^{\prime}\right)|V^{\prime}\in\mathbf{V}^{\mathcal{G}}\right\} $
and for each node (signature object) $V\in\mathbf{V}^{\mathcal{G}}$
with corresponding causal module (signature morphism),
\begin{equation}
\begin{aligned}\mathrm{dom}\left(\mathrm{Module}\left(V\right)\right) & =\bigotimes_{V^{\prime}\in\mathrm{pa}_{\mathcal{G}}\left(V\right)}V^{\prime}\\
\mathrm{cod}\left(\mathrm{Module}\left(V\right)\right) & =V^{\left|\mathrm{ch}_{\mathcal{G}}\left(V\right)\right|+1}.
\end{aligned}
\end{equation}

Our formulation of syntactic causal identification can only be applied
where the domain and codomain of the causal module $v=\mathrm{Module}\left(V\right)$
is explicit in a signature. Therefore, in practice, syntactic causal
identification requires the undirectional part of the ADMG to be available
in \emph{chain-factored }form. This is obtained using any topological
ordering of the ADMG as $\Sigma^{\mathcal{F}}=\left(\Sigma_{0}^{\mathcal{G}},\Sigma_{1}^{\mathcal{G}},\mathrm{dom},\mathrm{cod}\right)$
where,

\begin{equation}
\begin{aligned}\mathrm{dom}\left(\mathrm{Module}\left(V\right)\right) & =\bigotimes_{V^{\prime}\in\mathrm{pre}_{\mathcal{G}}\left(V\right)}V^{\prime}\\
\mathrm{cod}\left(\mathrm{Module}\left(V\right)\right) & =V^{\left|\mathrm{succ}_{\mathcal{G}}\left(V\right)\right|+1}.
\end{aligned}
\end{equation}

It is useful to have access to parent and child information from a
signature. The set of parent modules of causal module $v=\mathrm{Module}\left(V\right)$
are given by $\mathrm{pa}_{\Sigma}\left(V\right)=\left\{ v^{\prime}\in\Sigma_{1}:\mathrm{cod}\left(v^{\prime}\right)\cap\mathrm{dom}\left(\mathrm{Module}\left(V\right)\right)\ne\emptyset\right\} $
and child modules by $\mathrm{ch}_{\Sigma}\left(V\right)=\left\{ v^{\prime}\in\Sigma_{1}:\mathrm{dom}\left(v^{\prime}\right)\cap\mathrm{cod}\left(\mathrm{Module}\left(V\right)\right)\ne\emptyset\right\} $.

\subsection{Syntactic causal identification}

Here we present our main result. \citet[Theorem 49]{Richardson:2012}
is a re-formulation of the \emph{ID algorithm} \citep{Shpitser:2008}
for causal identification in general causal models with latent variables,
in terms of fixing\emph{ }operations on conditional ADMGs (CADMGs).
In this section we provide a purely syntactic description of the same
algorithm which uses only the structural information in the ADMG.
\begin{thm}
In the ADMG $\mathcal{G}$, consider the set of \emph{cause }$\mathbf{A}\subset\mathbf{V}^{\mathcal{G}}$
and effect variables $\mathbf{Y}\subset\mathbf{V}^{\mathcal{G}}$,
where $\mathbf{A}$ and $\mathbf{Y}$ do not intersect. Now consider
the set of variables $\mathbf{Y}^{\star}=\mathrm{an}_{\mathcal{G}_{\mathbf{V}^{\mathcal{G}}\backslash\mathbf{A}}}\left(\mathbf{Y}\right)$
and $\mathbf{D}^{\star}$ the set of districts of the subgraph $\mathcal{G}_{\mathbf{Y}^{\star}}$.
The signature of the \emph{syntactic causal effect}, $\Sigma_{\mathbf{Y}|\mathrm{do}\left(\mathbf{A}\right)}^{\mathcal{G}}$,
of $\mathbf{A}$ on $\mathbf{Y}$ is \emph{identifiable} if, for every
district $\mathbf{D}^{\prime}\in\mathbf{D}^{\star}$ the set of nodes
$\mathbf{V}^{\mathcal{G}}\backslash\mathbf{D}^{\prime}$ is a valid
fixing sequence. If identifiable, this causal effect is given by the
following composite signature manipulation,

\begin{equation}
\Sigma_{\mathbf{Y}|\mathrm{do}\left(\mathbf{A}\right)}^{\mathcal{G}}=\mathrm{Hide}_{\mathbf{Y}^{\star}\backslash\mathbf{Y}}\left(\bigcup_{\mathbf{D}^{\prime}\in\mathbf{D}^{\star}}\mathrm{Simple}\left(\mathrm{Fixseq}_{\mathbf{V}^{\mathcal{G}}\backslash\mathbf{D}^{\prime}}\left(\Sigma^{\mathcal{F}}\right)\right)\right).\label{eq:identify}
\end{equation}
\end{thm}

\subsection{Signature manipulations}

This section details the manipulations required to implement the syntactic
identification algorithm of (\ref{eq:identify}).
\begin{defn}
(Marginalization/hiding)For a chain-factored signature $\Sigma^{\mathcal{F}}=\left(\Sigma_{0}^{\mathcal{F}},\Sigma_{1}^{\mathcal{F}},\mathrm{dom},\mathrm{cod}\right)$,
the analogue of \emph{marginalization} of a single variable $V$ from
a distribution over a set of variables, is given by $\mathrm{Hide}_{V}\left(\Sigma^{\mathcal{F}}\right)=\left(\Sigma_{0}^{\mathcal{F}},\Sigma_{1}^{\mathcal{F}},\mathrm{dom},\mathrm{cod}^{\prime}\right)$,
where for $V$, 
\begin{equation}
\mathrm{cod}^{\prime}\left(\mathrm{Module}\left(V\right)\right)=V^{\left|\mathrm{succ}_{\mathcal{G}}\left(V\right)\right|}
\end{equation}
 and $\mathrm{cod}^{\prime}=\mathrm{cod}$ otherwise. For the set
$\mathbf{W}=\left\{ V_{1},\ldots,V_{k}\right\} \subset\mathbf{V}^{\mathcal{G}}$,
we extend this function to the composite $\mathrm{Hide}_{\mathbf{W}}=\mathrm{Hide}_{V_{1}}\circ\cdots\circ\mathrm{Hide}_{V_{k}}$.
\end{defn}
Pearl's causal interventions in ADMGs requires deleting parent edges
\citep{Pearl:2009,Richardson:2012,Bareinboim:2020}. For affine SMC
signatures, this entails replacing causal modules with identity/copy
morphisms and deleting any wires connected to that module. This is
captured in the following definition. 
\begin{defn}
(Causal control). Given a symmetric monoidal signature $\Sigma=\left(\Sigma_{0},\Sigma_{1},\text{dom},\text{cod}\right)$,
and an object $V\in\Sigma_{0}$. The function $\mathrm{Control}_{V}\left(\Sigma\right)=\left(\Sigma_{0},\Sigma_{1},\mathrm{dom}^{\prime},\mathrm{cod}^{\prime}\right),$manipulates
the signature in the following way
\begin{equation}
\mathrm{dom}^{\prime}\left(\mathrm{Module}\left(V\right)\right)=V,
\end{equation}
which replaces the module with a copied identity morphism, and for
all other $v^{\prime}\in\Sigma_{1}$ such that $v^{\prime}\ne\mathrm{Module}\left(V\right)$,
\begin{equation}
\begin{aligned}\mathrm{dom}^{\prime}\left(v^{\prime}\right) & =\mathrm{dom}\left(v^{\prime}\right)\\
\mathrm{cod}^{\prime}\left(v^{\prime}\right) & =\mathrm{cod}\left(v^{\prime}\right)\backslash\mathrm{dom}\left(\mathrm{Module}\left(V\right)\right),
\end{aligned}
\end{equation}
where the second line deletes incoming wires using the \emph{multiset
difference}. This operation is extended to controlling a set as with
marginalization above.
\end{defn}
\begin{defn}
(Causal fixing). Given a symmetric monoidal signature $\Sigma=\left(\Sigma_{0},\Sigma_{1},\text{dom},\text{cod}\right)$,
and an object $V\in\Sigma_{0}$. The syntactic fixing operation is
the composition of marginalization and control functions 
\begin{equation}
\mathrm{Fix}_{V}=\mathrm{Control}_{V}\circ\mathrm{Hide}_{V},
\end{equation}
\end{defn}
The above definition is exactly the syntactic analogue of fixing in
ADMGs \citep{Richardson:2012}.

To ensure identifiability, fixing can only be applied to objects which
are fixable relative to some signature, $\Sigma^{\mathcal{G}}$ derived
from an ADMG $\mathcal{G}$. Given a set of objects $\mathbf{W}=\left\{ V_{1},\dots,V_{k}\right\} $
to fix, we need to determine a valid sequence of fixing operations,
$\mathrm{Fixseq}_{\mathbf{W}}$, for this set \citep{Richardson:2012}.
This is computed recursively as follows. To initialize the recursion,
set $\mathbf{W}^{\prime}=\mathbf{W}$, initialize $\Sigma=\Sigma^{\mathcal{G}}$,
initialize the fixing sequence operation $\mathrm{Fixseq}_{\mathbf{W}}=\mathit{id}$
(the identity operation). The recursion step is as follows: choose
any $V\in\mathbf{W}^{\prime}$ such that $V$ is fixable in $\Sigma$.
If there exists no such $V$ then the sequence $\mathbf{W}$ cannot
have a valid fixing sequence and the recursion terminates. Otherwise,
if $\mathrm{ch}_{\Sigma}\left(V\right)=\emptyset$ then update $\mathrm{Fixseq}_{\mathbf{W}}\mapsto\mathrm{Hide}_{V}\circ\mathrm{Fixseq}_{\mathbf{W}}$,
otherwise update $\mathrm{Fixseq}_{\mathbf{W}}\mapsto\mathrm{Fix}_{V}\circ\mathrm{Fixseq}_{\mathbf{W}}$
instead. Now, apply this operation to obtain the updated signature
$\Sigma\mapsto\mathrm{Fixseq}_{\mathbf{W}}\left(\Sigma^{\mathcal{G}}\right)$
and delete $V$ from the fixing set, $\mathbf{W}^{\prime}\mapsto\mathbf{W}^{\prime}\backslash V$.
If all objects to fix have been exhausted, i.e. $\mathbf{W}^{\prime}=\emptyset$,
then the recursion terminates with the fixing operation sequence $\mathrm{Fixseq}_{\mathbf{W}}$,
otherwise the process returns to the recursion step.

Manipulating a signature can lead to a causal module $v=\mathrm{Module}\left(V\right)$
being equivalent to the identity morphism i.e. where $\mathrm{dom}\left(\mathrm{Module}\left(V\right)\right)=\mathrm{cod}\left(\mathrm{Module}\left(V\right)\right)=V$.
These can be deleted from the signature, simplifying the causal model
it specifies. Writing the set of modules which are not equivalent
to a identity as $\mathbf{W}=\left\{ V\in\Sigma_{0}^{\mathcal{G}}|\left(\mathrm{dom}\left(\mathrm{Module}\left(V\right)\right)\ne V\right)\vee\left(\mathrm{cod}\left(\mathrm{Module}\left(V\right)\right)\ne V\right)\right\} $,
then $\mathrm{DeleteId}\left(\Sigma\right)=\left(\mathbf{W},\left\{ \mathrm{Module}\left(V^{\prime}\right)|V^{\prime}\in\mathbf{W}\right\} ,\mathrm{dom},\mathrm{cod}\right)$
is the simplified signature. Furthermore, signature manipulation can
lead to a causal module having no downstream effects. Such modules
can also be deleted from the signature, further simplifying the causal
model.
\begin{defn}
(Signature simplification). Consider the set $\mathbf{W}=\left\{ V\in\Sigma_{0}|\mathrm{cod}\left(\mathrm{Module}\left(V\right)\right)\ne1\right\} $
of objects whose causal modules do not have marginalized outputs,
then the simplified signature can be written,
\begin{equation}
\mathrm{Simplify}\left(\Sigma\right)=\left(\mathbf{W},\left\{ \mathrm{Module}\left(V^{\prime}\right)|V^{\prime}\in\mathbf{W}\right\} ,\mathrm{dom},\mathrm{cod}^{\prime}\right)\label{eq:simplify}
\end{equation}
with modified codomain $\mathrm{cod}^{\prime}\left(\mathrm{Module}\left(V^{\prime}\right)\right)=\mathrm{cod}\left(\mathrm{Module}\left(V^{\prime}\right)\right)\backslash\mathrm{dom}\left(\mathrm{Module}\left(W^{\prime}\right)\right)$
for all $V\in\Sigma_{0}\backslash\mathbf{W}$ and $W^{\prime}\in\mathbf{W}$.
\end{defn}
Deleting a module might lead to other causal modules having no downstream
effects, therefore it is necessary to iterate (\ref{eq:simplify})
until a fixed point signature is reached. Formally, starting at $\Sigma^{0}=\Sigma$,
the iteration $\Sigma^{n+1}=\mathrm{Simplify}\left(\Sigma^{n}\right)$
is repeated until some $N$ is obtained such that $\mathrm{Simplify}\left(\Sigma^{N}\right)=\Sigma^{N}$,
whereupon we use $\mathrm{Simple}\left(\Sigma\right)=\mathrm{DeleteId}\left(\Sigma^{N}\right)$
as the fully simplified signature.
\begin{defn}
(Combining exterior signatures). For signatures $\Sigma^{1}$ and
$\Sigma^{2}$, their combination is,
\begin{equation}
\begin{aligned}\Sigma & =\mathrm{Ext}\left(\Sigma^{1}\right)\cup\mathrm{Ext}\left(\Sigma^{2}\right)\\
 & =\left(\Sigma_{0}^{1}\cup\Sigma_{0}^{2},\Sigma_{1}^{1}\cup\Sigma_{1}^{2},\mathrm{dom}^{1}\cup\mathrm{dom}^{2},\mathrm{cod}^{\prime}\right)
\end{aligned}
\end{equation}
where $\mathrm{Ext}\left(\Sigma^{1}\right)=\left(\Sigma_{0}^{1},\Sigma_{1}^{1},\mathrm{dom}^{1},\mathrm{cod}^{1}\right)$,
$\mathrm{Ext}\left(\Sigma^{2}\right)=\left(\Sigma_{0}^{2},\Sigma_{1}^{2},\mathrm{dom}^{2},\mathrm{cod}^{2}\right)$
are the signatures of the exteriors of $\Sigma^{1}$ and $\Sigma^{2}$
respectively, and their combined codomain is,
\begin{align}
\mathrm{cod}^{\prime}\left(\mathrm{Module}\left(V\right)\right) & =V^{\left|\mathrm{ch}_{\Sigma}\left(V\right)\right|+1}
\end{align}
for all $V\in\Sigma_{0}^{1}\cup\Sigma_{0}^{2}$ for which a causal
module is assigned through $\mathrm{cod}^{1}$ and $\mathrm{cod}^{2}$.
\end{defn}

\section{Applications}

\subsection{Back-door adjustment: simple case}

As first application, we show how to derive a purely syntactic account
of back-door adjustment. Consider a simple, fully observed model with
one confounder. The ADMG is defined by $\mathbf{V}^{\mathcal{G}}=\left\{ X,Y,U\right\} $
and edges $\mathbf{E}^{\mathcal{G}}=\left\{ X\to Y,U\to X,U\to Y\right\} $,
and we want the interventional signature $\Sigma_{Y|\mathrm{do}\left(X\right)}^{\mathcal{G}}$.
In this situation, the model is equivalent to a fully chain-factored
graph so we can directly fix $X$ in the signature without the need
for explicit district decomposition.

The corresponding signature $\Sigma^{\mathcal{G}}=\left(\left\{ X,Y,U\right\} ,\left\{ x,y,u\right\} ,\left\{ u:1\to U^{3},x:U\to X^{2},y:XU\to Y\right\} \right)$
so that, on application of fixing, we obtain the manipulated signature,

\begin{equation}
\begin{aligned}\Sigma & =\mathrm{Fix}_{X}\left(\left\{ X,Y,U\right\} ,\left\{ x,y,u\right\} ,\left\{ u:1\to U^{3},x:U\to X^{2},y:XU\to Y\right\} \right)\\
 & =\left(\mathrm{Control}_{X}\circ\mathrm{Hide}_{X}\right)\left(\left\{ X,Y,U\right\} ,\left\{ x,y,u\right\} ,\left\{ u:1\to U^{3},x:U\to X^{2},y:XU\to Y\right\} \right)\\
 & =\left(\mathrm{Control}_{X}\right)\left(\left\{ X,Y,U\right\} ,\left\{ x,y,u\right\} ,\left\{ u:1\to U^{3},x:U\to X,y:XU\to Y\right\} \right)\\
 & =\left(\left\{ X,Y,U\right\} ,\left\{ x,y,u\right\} ,\left\{ u:1\to U^{2},x:X\to X,y:XU\to Y\right\} \right),
\end{aligned}
\end{equation}
which, when simplified becomes,

\begin{equation}
\begin{aligned}\mathrm{Simple}\left(\Sigma\right) & =\mathrm{Simple}\left(\left\{ X,Y,U\right\} ,\left\{ x,y,u\right\} ,\left\{ u:1\to U^{2},x:X\to X,y:XU\to Y\right\} \right)\\
 & =\left(\left\{ X,Y,U\right\} ,\left\{ y,u\right\} ,\left\{ u:1\to U^{2},y:XU\to Y\right\} \right)\\
 & =\Sigma_{YU|\mathrm{do}\left(X\right)}^{\mathcal{G}},
\end{aligned}
\end{equation}
and then finally marginalizing out $U$ we obtain the desired interventional
signature,

\begin{equation}
\begin{aligned}\Sigma_{Y|\mathrm{do}\left(X\right)}^{\mathcal{G}} & =\mathrm{Hide}_{U}\left(\left\{ X,Y,U\right\} ,\left\{ y,u\right\} ,\left\{ u:1\to U^{2},y:XU\to Y\right\} \right)\\
 & =\left(\left\{ X,Y,U\right\} ,\left\{ y,u\right\} ,\left\{ u:1\to U,y:XU\to Y\right\} \right).
\end{aligned}
\label{eq:back-door}
\end{equation}

This is the purely syntactic categorical analogue of the well-known
\emph{back-door adjustment formula} in this simple case. When this
signature is interpreted as a Markov category \citep{Fritz:2020}
with continuous sample spaces $X\mapsto\Omega_{X}$, $U\mapsto\Omega_{U}$
and $Y\mapsto\Omega_{Y}$ with composition $\cdot$ corresponding
to the \emph{Chapman-Kolmogorov equation }\citep{little2019machine,Jacobs:2021}
and the causal morphisms are \emph{conditional distributions} $u\mapsto p\left(U\right)$
and $y\mapsto p\left(Y|X,U\right)$, then (\ref{eq:back-door}) is
the interventional distribution,

\begin{equation}
p\left(Y=y|\mathrm{do}\left(X=x\right)\right)=\int_{\Omega_{U}}p\left(Y=y|X=x,U=u\right)p\left(U=u\right)du.
\end{equation}

Another interesting interpretation is in terms of the \emph{min-plus
semifield Markov category }\citep{Cakiqi:2022} for which
\begin{equation}
q\left(y|\mathrm{do}\left(x\right)\right)=\min_{u\in U}\left[q\left(y|x,u\right)+q\left(u\right)\right],
\end{equation}
where $u\mapsto q\left(u\right)$ is an \emph{inferential bias }and
$y\mapsto q\left(y|x,u\right)$ a \emph{clique potential }widely encountered
in machine learning. For these functions to be biases/potentials,
they must be \emph{normalizable} in the min-plus semifield, i.e. $\min_{u\in U}q\left(u\right)=0$
and $\min_{y\in Y}q\left(y|x,u\right)=0$ \citep{little2019machine}.

\subsection{Front-door adjustment}

\begin{figure}
\begin{centering}
\begin{tabular}{ll>{\raggedright}p{2cm}lllll}
\begin{tikzpicture}[unit length/.code={{\newdimen\tikzunit}\setlength{\tikzunit}{#1}},unit length=8mm,x=\tikzunit,y=\tikzunit,semithick,box/.style={rectangle,draw,solid,rounded corners},junction/.style={circle,draw,fill,inner sep=0},outer box/.style={draw=none},wire/.style={draw,postaction={decorate},decoration={markings,mark=at position 0.5 with {#1}}}] 
\node[box,minimum size=1\tikzunit] (x) at (0,2) {$x$}; 
\node[box,minimum size=1\tikzunit] (z) at (0,4) {$z$}; 
\node (a) at (0,0) {(a)}; 
\node (Y) at (0,7.5) {$Y$};
\node (X) at (-1.5,7.5) {$X$};
\node (Z) at (-1,7.5) {$Z$};
\node[box,minimum height=1\tikzunit,minimum width=1\tikzunit] (y) at (0,6) {$y$}; 
\node[junction,minimum size=0.25\tikzunit] (xcopy) at (0,3) {}; 
\node[junction,minimum size=0.25\tikzunit] (zcopy) at (0,5) {}; 
\path[wire] (x.north) to[out=90, in=-90] (xcopy.south); 
\path[wire] (z.north) to[out=90, in=-90] (zcopy.south); 
\path[wire] (xcopy.east) to[out=180, in=-90] (z.south); 
\path[wire] (zcopy.east) [out=180, in=-90] to ($(y.south)-(0,0)$);
\path[wire] (xcopy.east) to[out=180,in=-90] (-1.5,4) to[out=90,in=-90] (X);
\path[wire] (zcopy.east) to[out=180,in=-90] (-1,6) to[out=90,in=-90] (Z);
\path[wire] (y.north) to (Y); 
\end{tikzpicture} & \begin{tikzpicture}[unit length/.code={{\newdimen\tikzunit}\setlength{\tikzunit}{#1}},unit length=8mm,x=\tikzunit,y=\tikzunit,semithick,box/.style={rectangle,draw,solid,rounded corners},junction/.style={circle,draw,fill,inner sep=0},outer box/.style={draw=none},wire/.style={draw,postaction={decorate},decoration={markings,mark=at position 0.5 with {#1}}}]
\node[box,minimum size=1\tikzunit] (x) at (0,2) {$x$};
\node (a) at (0,0) {(b)}; 
\node[box,minimum height=1\tikzunit,minimum width=2\tikzunit] (x3) at (1,6) {$y$};
\node[box,minimum size=1\tikzunit] (x4) at (0,4) {$z$};
\node[junction,minimum size=0.25\tikzunit] (xcopy) at (0,3) {};
\node[junction,minimum size=0.25\tikzunit] (x4copy) at (0,5) {};
\node (Y) at (1,7.5) {$Y$};
\node (X) at (-1,7.5) {$X$};
\node (Z) at (-0.5,7.5) {$Z$};
\path[wire] (x.north) to[out=90, in=-90] (xcopy.south);
\path[wire] (x4.north) to[out=90, in=-90] (x4copy.south);
\path[wire] (xcopy.east) to[out=180, in=-90]  (x4.south);
\path[wire] (x4copy.east)  to[out=0,in=-90] (0.5,5.5) to[out=90, in=-90] ($(x3.south)-(0.5,0)$);
\path[wire] (xcopy.east) to[out=180,in=-90] (-1,4) to[out=90,in=-90] (X);
\path[wire] (xcopy.east)  to[out=0, in=-90] (1.5,4) to[out=90, in=-90] ($(x3.south)+(0.5,0)$);
\path[wire] (x4copy.east) to[out=180,in=-90] (-0.5,6) to[out=90,in=-90] (Z);
\path[wire] (x3.north) to (Y);
\end{tikzpicture} & \begin{tikzpicture}[unit length/.code={{\newdimen\tikzunit}\setlength{\tikzunit}{#1}},unit length=8mm,x=\tikzunit,y=\tikzunit,semithick,box/.style={rectangle,draw,solid,rounded corners},junction/.style={circle,draw,fill,inner sep=0},outer box/.style={draw=none},wire/.style={draw,postaction={decorate},decoration={markings,mark=at position 0.5 with {#1}}}]
\node[box,minimum size=1\tikzunit] (x) at (2.5,4) {$x$};
\node[box,minimum height=1\tikzunit,minimum width=2\tikzunit] (x3) at (2,6) {$y$};
\node (Y) at (2,9.5) {$Y$};
\node (a) at (2,2) {(c)}; 
\node (x4) at (1.5,3.7) {$Z$};
\path[wire] (x.north) to[out=90, in=-90] ($(x3.south)+(0.5,0)$);
\path[wire] (x4.north) to[out=90, in=-90] ($(x3.south)-(0.5,0)$);
\path[wire] (x3.north) to[out=90, in=-90] (Y);
\end{tikzpicture} & \begin{tikzpicture}[unit length/.code={{\newdimen\tikzunit}\setlength{\tikzunit}{#1}},unit length=8mm,x=\tikzunit,y=\tikzunit,semithick,box/.style={rectangle,draw,solid,rounded corners},junction/.style={circle,draw,fill,inner sep=0},outer box/.style={draw=none},wire/.style={draw,postaction={decorate},decoration={markings,mark=at position 0.5 with {#1}}}]
\node[box,minimum height=1\tikzunit,minimum width=1\tikzunit] (x3) at (2,1.2) {$q$};
\node (x4) at (2,-1) {$Z$};
\node (Y) at (2,4.7) {$Y$};
\node (a) at (2,-2.75) {(d)}; 
\path[wire] (x4.north) to[out=90, in=-90] ($(x3.south)-(00,0)$);
\path[wire] (x3.north) to[out=90, in=-90] (Y) ;
\end{tikzpicture} & \begin{tikzpicture}[unit length/.code={{\newdimen\tikzunit}\setlength{\tikzunit}{#1}},unit length=8mm,x=\tikzunit,y=\tikzunit,semithick,box/.style={rectangle,draw,solid,rounded corners},junction/.style={circle,draw,fill,inner sep=0},outer box/.style={draw=none},wire/.style={draw,postaction={decorate},decoration={markings,mark=at position 0.5 with {#1}}}]
\node (x) at (0,-1) {$X$};
\node (Z) at (0,4.7) {$Z$};
\node[box,minimum size=1\tikzunit] (x4) at (0,1.2) {$z$};
\node (a) at (0,-2.75) {(e)}; 
\path[wire] (x) to[out=90, in=-90] (x4.south);
\path[wire] (x4.north) to[out=90, in=-90] (Z);
\end{tikzpicture} & \begin{tikzpicture}[unit length/.code={{\newdimen\tikzunit}\setlength{\tikzunit}{#1}},unit length=8mm,x=\tikzunit,y=\tikzunit,semithick,box/.style={rectangle,draw,solid,rounded corners},junction/.style={circle,draw,fill,inner sep=0},outer box/.style={draw=none},wire/.style={draw,postaction={decorate},decoration={markings,mark=at position 0.5 with {#1}}}]
\node (X) at (0,2) {$X$};
\node (a) at (0,0.25) {(f)}; 
\node (Y) at (0,7.7) {$Y$};
\node (Z) at (-1,7.7) {$Z$};
\node[box,minimum height=1\tikzunit,minimum width=1\tikzunit] (x3) at (0,6.2) {$q$};
\node[box,minimum size=1\tikzunit] (x4) at (0,4.2) {$z$};
\node[junction,minimum size=0.25\tikzunit] (x4copy) at (0,5.2) {};
\path[wire] (X.north) to[out=90, in=-90] (x4.south);
\path[wire] (x4.north) to[out=90, in=-90] (x4copy.south);
\path[wire] (x4copy.east) to[out=180, in=-90] ($(x3.south)-(0,0)$);
\path[wire] (x4copy.east)  to[out=180,in=-90] (-1,6) to[out=90,in=-90] (Z);
\path[wire] (x3.north) to[out=90, in=-90] (Y);
\end{tikzpicture} & \begin{tikzpicture}[unit length/.code={{\newdimen\tikzunit}\setlength{\tikzunit}{#1}},unit length=8mm,x=\tikzunit,y=\tikzunit,semithick,box/.style={rectangle,draw,solid,rounded corners},junction/.style={circle,draw,fill,inner sep=0},outer box/.style={draw=none},wire/.style={draw,postaction={decorate},decoration={markings,mark=at position 0.5 with {#1}}}]
\node (X) at (2,2) {$X$};
\node (Z) at (2.25,5.2) {$Z$};
\node (Y) at (2,7.7) {$Y$};
\node (a) at (2,0.25) {(g)}; 
\node[box,minimum height=1\tikzunit,minimum width=1\tikzunit] (x3) at (2,6.2) {$q$};
\node[box,minimum size=1\tikzunit] (x4) at (2,4.2) {$z$};
\path[wire] (X.north) to[out=90, in=-90] (x4.south);
\path[wire] (x4.north) to[out=90, in=-90] ($(x3.south)-(0,0)$);
\path[wire] (x3.north) to[out=90, in=-90] (Y);
\end{tikzpicture} & \begin{tikzpicture}[unit length/.code={{\newdimen\tikzunit}\setlength{\tikzunit}{#1}},unit length=8mm,x=\tikzunit,y=\tikzunit,semithick,box/.style={rectangle,draw,solid,rounded corners},junction/.style={circle,draw,fill,inner sep=0},outer box/.style={draw=none},wire/.style={draw,postaction={decorate},decoration={markings,mark=at position 0.5 with {#1}}}]
\node[box,minimum size=1\tikzunit] (xx) at (2.75,4.69) {$x^{\prime}$};
\node[box,minimum height=1\tikzunit,minimum width=2\tikzunit] (x3) at (2,6.69) {$y$};
\node (x) at (1.25,2.5) {$X$};
\node (Y) at (2,8.25) {$Y$};
\node (Z) at (1.5,5.69) {$Z$};
\node (XX) at (3.1,5.69) {$X^{\prime}$};
\node (a) at (2,0.78) {(h)}; 
\node[box,minimum size=1\tikzunit] (x4) at (1.25,4.69) {$z$};
\path[wire] (x) to[out=90, in=-90] (x4.south);
\path[wire] (xx.north) to[out=90, in=-90] ($(x3.south)+(0.75,0)$);
\path[wire] (x4.north) to[out=90, in=-90] ($(x3.south)-(0.75,0)$);
\path[wire] (x3.north) to[out=90, in=-90] (Y);
\end{tikzpicture}\tabularnewline
\end{tabular}
\par\end{centering}
\caption{String diagram representations of maximal causal models for monoidal
signatures obtained during the derivation of the purely syntactic
front-door adjustment interventional signature $\Sigma_{Y|\mathrm{do}\left(X\right)}^{\mathcal{G}}$.
Signatures are as follows: (a) $\Sigma^{\mathcal{G}}$(front-door
ADMG, observable variables only); (b) $\Sigma^{\mathcal{F}}$ (chain-factored
observable ADMG); (c) $\Sigma^{1}=\left(\mathrm{Simple}\circ\mathrm{Fixseq}_{\left\{ X,Z\right\} }\right)\left(\Sigma^{\mathcal{F}}\right)$
(fixed district $\left\{ Y\right\} $); (d) $\mathrm{Ext}\left(\Sigma^{1}\right)$
(exterior signature of $\Sigma^{1}$); (e) $\Sigma^{2}=\left(\mathrm{Simple}\circ\mathrm{Fixseq}_{\left\{ X,Y\right\} }\right)\left(\Sigma^{\mathcal{F}}\right)$
(fixed district $\left\{ Z\right\} $); (f) $\Sigma_{ZY|\mathrm{do}\left(X\right)}^{\mathcal{G}}=\mathrm{Ext}\left(\Sigma^{1}\right)\cup\mathrm{Ext}\left(\Sigma^{2}\right)$
(interventional signature); (g) $\Sigma_{Y|\mathrm{do}\left(X\right)}^{\mathcal{G}}$
(marginalized interventional signature) and (h) $\Sigma_{Y|\mathrm{do}\left(X\right)}^{\mathcal{G}}$
with internal mechanisms made explicit.\label{fig:front-door}}

\end{figure}
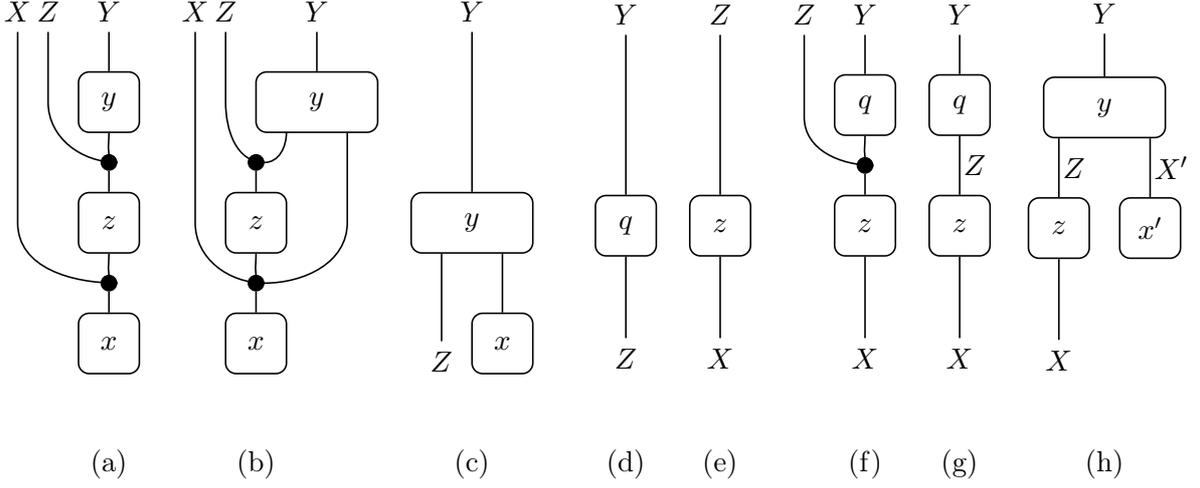
The front-door graph \citep{Pearl:2009} is an ADMG $\mathcal{G}$
given by $\mathbf{V}^{\mathcal{G}}=\left\{ X,Y,Z\right\} $ with edges
$\mathbf{E}^{\mathcal{G}}=\left\{ X\to Z\to Y,X\leftrightarrow Y\right\} $.
We want the pure syntactic interventional signature $\Sigma_{Y|\mathrm{do}\left(X\right)}^{\mathcal{G}}$,
which represents the causal effect on $\mathbf{Y}=\left\{ Y\right\} $
of $\mathbf{A}=\left\{ X\right\} $. Here, $\mathbf{V}^{\mathcal{G}}\backslash\mathbf{A}=\left\{ Y,Z\right\} $
so that $\mathbf{Y}^{\star}=\mathrm{an}_{\mathcal{G}_{\left\{ Y,Z\right\} }}\left(\left\{ Y\right\} \right)=\left\{ Y,Z\right\} $
with corresponding districts $\mathbf{D}^{\star}=\left\{ \left\{ Y\right\} ,\left\{ Z\right\} \right\} $
of subgraph $\mathcal{G}_{\left\{ Y,Z\right\} }$ (See Figure \ref{fig:front-door}).The
monoidal signature of $\mathcal{G}$ is,

\begin{equation}
\Sigma^{\mathcal{G}}=\left(\left\{ X,Y,Z\right\} ,\left\{ x,y,z\right\} ,\left\{ x:1\to X^{2},z:X\to Z^{2},y:Z\to Y\right\} \right),
\end{equation}
with corresponding chain-factored signature,

\begin{equation}
\Sigma^{\mathcal{F}}=\left(\left\{ X,Y,Z\right\} ,\left\{ x,y,z\right\} ,\left\{ x:1\to X^{3},z:X\to Z^{2},y:XZ\to Y\right\} \right).
\end{equation}

For the district $\mathbf{D}^{\prime}=\left\{ Y\right\} $, the fixing
set is $\mathbf{W}=\left\{ X,Z\right\} $, for which we obtain the
sequence $\mathrm{Fixseq}_{\left\{ X,Z\right\} }=\mathrm{Hide}_{X}\circ\mathrm{Fix}_{Z}$
applied to $\Sigma^{\mathcal{F}}$,

\begin{equation}
\begin{aligned}\Sigma^{1} & =\left(\mathrm{Simple}\circ\mathrm{Hide}_{X}\circ\mathrm{Fix}_{Z}\right)\left(\Sigma^{\mathcal{F}}\right)\\
 & =\left(\mathrm{Simple}\circ\mathrm{\mathrm{Hide}}_{X}\circ\mathrm{Fix}_{Z}\right)\left(\left\{ X,Y,Z\right\} ,\left\{ x,y,z\right\} ,\left\{ x:1\to X^{3},z:X\to Z^{2},y:XZ\to Y\right\} \right)\\
 & =\left(\mathrm{Simple}\circ\mathrm{\mathrm{Hide}}_{X}\right)\left(\left\{ X,Y,Z\right\} ,\left\{ x,y,z\right\} ,\left\{ x:1\to X^{2},z:Z\to Z,y:XZ\to Y\right\} \right)\\
 & =\mathrm{Simple}\left(\left\{ X,Y,Z\right\} ,\left\{ x,y,z\right\} ,\left\{ x:1\to X,z:Z\to Z,y:XZ\to Y\right\} \right)\\
 & =\left(\left\{ X,Y,Z\right\} ,\left\{ x,y\right\} ,\left\{ x:1\to X,y:XZ\to Y\right\} \right),
\end{aligned}
\end{equation}
so that $\mathrm{Ext}\left(\Sigma^{1}\right)=\left(\left\{ Y,Z\right\} ,\left\{ q\right\} ,\left\{ q:Z\to Y\right\} \right)$
with $q=y\cdot\left(\mathit{id}_{Z}\otimes x\right)$ and $\mathrm{Module}\left(Y\right)=q$.

For the district $\mathbf{D}^{\prime}=\left\{ Z\right\} $, the fixing
set is $\mathbf{W}=\left\{ X,Y\right\} $, for which we obtain the
sequence $\mathrm{Fixseq}_{\left\{ X,Y\right\} }=\mathrm{Fix}_{X}\circ\mathrm{Fix}_{Y}$,

\begin{equation}
\begin{aligned}\Sigma^{2} & =\left(\mathrm{Simple}\circ\mathrm{Fix}_{X}\circ\mathrm{Fix}_{Y}\right)\left(\Sigma^{\mathcal{F}}\right)\\
 & =\left(\mathrm{Simple}\circ\mathrm{Fix}_{X}\circ\mathrm{Fix}_{Y}\right)\left(\left\{ X,Y,Z\right\} ,\left\{ x,y,z\right\} ,\left\{ x:1\to X^{3},z:X\to Z^{2},y:XZ\to Y\right\} \right)\\
 & =\left(\mathrm{Simple}\circ\mathrm{Fix}_{X}\right)\left(\left\{ X,Y,Z\right\} ,\left\{ x,y,z\right\} ,\left\{ x:1\to X^{2},z:X\to Z,y:Y\to1\right\} \right)\\
 & =\mathrm{Simple}\left(\left\{ X,Y,Z\right\} ,\left\{ x,y,z\right\} ,\left\{ x:X\to X,z:X\to Z,y:Y\to1\right\} \right)\\
 & =\left(\left\{ X,Z\right\} ,\left\{ z\right\} ,\left\{ z:X\to Z\right\} \right)\\
 & =\mathrm{Ext}\left(\Sigma^{2}\right),
\end{aligned}
\end{equation}
with $\mathrm{Module}\left(Z\right)=z$, where this signature is already
exterior because $z$ is not composite. Combining these two exterior
signatures, we obtain,

\begin{equation}
\begin{aligned}\Sigma_{ZY|\mathrm{do}\left(X\right)}^{\mathcal{G}} & =\mathrm{Ext}\left(\Sigma^{1}\right)\cup\mathrm{Ext}\left(\Sigma^{2}\right)\\
 & =\left(\left\{ X,Y,Z\right\} ,\left\{ q,z\right\} ,\left\{ z:X\to Z^{2},q:Z\to Y\right\} \right),
\end{aligned}
\end{equation}
and since $\mathbf{Y}^{\star}\backslash\mathbf{Y}=\left\{ Y,Z\right\} \backslash\left\{ Y\right\} =\left\{ Z\right\} $,
we obtain the desired interventional distribution by marginalization,

\begin{align*}
\Sigma_{Y|\mathrm{do}\left(X\right)}^{\mathcal{G}} & =\mathrm{Hide}_{\left\{ Z\right\} }\left(\Sigma_{ZY|\mathrm{do}\left(X\right)}^{\mathcal{G}}\right)\\
 & =\left(\left\{ X,Y,Z\right\} ,\left\{ q,z\right\} ,\left\{ z:X\to Z,q:Z\to Y\right\} \right).
\end{align*}

In practice, it may be useful to expose the interior of $q$ and to
do this we will need to relabel $x\to x^{\prime}$(and correspondingly,
$X\mapsto X^{\prime}$) inside $q$ to avoid a naming clash with the
interventional input $X$,
\begin{align}
\Sigma_{Y|\mathrm{do}\left(X\right)}^{\mathcal{G}} & =\left(\left\{ X,X^{\prime},Y,Z\right\} ,\left\{ x,y,z\right\} ,\left\{ x^{\prime}:1\to X^{\prime},z:X\to Z,y:X^{\prime}Z\to Y\right\} \right).\label{eq:front-door}
\end{align}

As with the back-door model, this is the purely syntactic categorical
analogue of the \emph{front-door adjustment formula}. As an example
interpretation, consider the Markov category with discrete sample
spaces $X^{\prime}\mapsto\Omega_{X}$, $Z\mapsto\Omega_{Y}$, $Y\mapsto\Omega_{Y}$
and with conditional distributions $x^{\prime}\mapsto p\left(X^{\prime}\right)$,
$z\mapsto p\left(Z|X\right)$ and $y\mapsto p\left(Y|X,Z\right)$,
then (\ref{eq:front-door}) is the familiar discrete interventional
distribution \citep{Pearl:2009},

\begin{align}
p\left(Y=y|\mathrm{do}\left(X=x\right)\right) & =\sum_{z\in\Omega_{Z}}p\left(Z=z|X=x\right)\sum_{x^{\prime}\in\Omega_{X}}p\left(Y=y|X^{\prime}=x^{\prime},Z=z\right)p\left(X^{\prime}=x^{\prime}\right).
\end{align}

Another useful interpretation are \emph{deterministic causal models
}in the SMC of sets and functions, in which composition $\cdot$ is
ordinary function composition and $\otimes$ is the pairing (bifunctor),
with the identity $1$ corresponding to the empty pair $\left(\,\right)$.
Then, interpreting $X^{\prime}$ with the set of possible values for
the constant $f_{X^{\prime}}:1\to X^{\prime}$, the functions $f_{Z}:X\to Z$
and $f_{Y}:X^{\prime}Z\to Y$, the front-door interventional model
corresponding to the signature $\Sigma_{Y|\mathrm{do}\left(X\right)}^{\mathcal{G}}$
is

\begin{equation}
\begin{aligned}f_{Y|\mathrm{do}\left(X\right)}\left(x\right) & =\pi_{2}\left(f_{Z}\left(x\right),f_{Y}\left(f_{X^{\prime}}\left(\,\right),f_{Z}\left(x\right)\right)\right)\\
 & =f_{Y}\left(f_{X^{\prime}}\left(\,\right),f_{Z}\left(x\right)\right),
\end{aligned}
\end{equation}
where $\pi_{2}$ is projection onto the second item of a pair.

\subsection{A more complex example}

\citet[Example 51]{Richardson:2012} describe an application of their
fixing theorem to a more complex causal model with four variables
and a single bidirected edge whose latent projection ADMG $\mathcal{G}$
is given by $\mathbf{V}^{\mathcal{G}}=\left\{ X_{1},X_{2},X_{3},X_{4}\right\} $
and by edge set
\begin{equation}
\mathbf{E}^{\mathcal{G}}=\left\{ X_{3}\leftarrow X_{1}\to X_{2},X_{2}\to X_{3},X_{3}\to X_{4},X_{2}\leftrightarrow X_{4}\right\} .
\end{equation}

In this example they identify the interventional distribution $p\left(X_{4}|\mathrm{do}\left(X_{2}\right)\right)$.
To illustrate our syntactic fixing algorithm for this example, we
want the monoidal signature of the pure syntactic causal effect on
$\mathbf{Y}=\left\{ X_{4}\right\} $ of $\mathbf{A}=\left\{ X_{2}\right\} $,
i.e. $\Sigma_{X_{4}|\mathrm{do}\left(X_{2}\right)}^{\mathcal{G}}$.
Here, $\mathbf{V}^{\mathcal{G}}\backslash\mathbf{A}=\left\{ X_{1},X_{3},X_{4}\right\} $
so that $\mathbf{Y}^{\star}=\mathrm{an}_{\mathcal{G}_{\mathbf{V}^{\mathcal{G}}\backslash\mathbf{A}}}\left(\left\{ X_{4}\right\} \right)=\left\{ X_{1},X_{3},X_{4}\right\} $
with corresponding subgraph districts $\mathbf{D}^{\star}=\mathcal{G}_{\mathbf{Y}^{\star}}=\left\{ \left\{ X_{1}\right\} ,\left\{ X_{3}\right\} ,\left\{ X_{4}\right\} \right\} $.
The monoidal signature of $\mathcal{G}$ is,
\begin{equation}
\begin{aligned}\Sigma_{0}^{\mathcal{G}} & =\left\{ X_{1},X_{2},X_{3},X_{4}\right\} \\
\Sigma_{1}^{\mathcal{G}} & =\left\{ x_{1},x_{2},x_{3},x_{4}\right\} \\
\mathrm{dom},\mathrm{cod} & =\left\{ x_{1}:1\to X_{1}^{3},x_{2}:X_{1}\to X_{2}^{2},x_{3}:X_{1}X_{2}\to X_{3}^{2},x_{4}:X_{1}X_{2}X_{3}\to X_{4}\right\} ,
\end{aligned}
\end{equation}
with corresponding chain-factored signature,
\begin{equation}
\begin{aligned}\Sigma_{0}^{\mathcal{F}} & =\left\{ X_{1},X_{2},X_{3},X_{4}\right\} \\
\Sigma_{1}^{\mathcal{F}} & =\left\{ x_{1},x_{2},x_{3},x_{4}\right\} \\
\mathrm{dom},\mathrm{cod} & =\left\{ x_{1}:1\to X_{1}^{4},x_{2}:X_{1}\to X_{2}^{3},x_{3}:X_{1}X_{2}\to X_{3}^{2},x_{4}:X_{1}X_{2}X_{3}\to X_{4}\right\} .
\end{aligned}
\end{equation}

For the district $\mathbf{D}^{\prime}=\left\{ X_{1}\right\} $, the
fixing set is $\mathbf{W}=\left\{ X_{2},X_{3},X_{4}\right\} $, for
which we obtain,

\begin{equation}
\begin{aligned}\Sigma^{1} & =\left(\mathrm{Simple}\circ\mathrm{Fix}_{X_{2}}\circ\mathrm{Fix}_{X_{3}}\circ\mathrm{Fix}_{X_{4}}\right)\left(\Sigma^{\mathcal{F}}\right)\\
 & =\left(\left\{ X_{1}\right\} ,\left\{ x_{1}\right\} ,\left\{ x_{1}:1\to X_{1}\right\} \right)\\
 & =\mathrm{Ext}\left(\Sigma^{1}\right)\:\mathrm{Module}\left(X_{1}\right)=x_{1}.
\end{aligned}
\end{equation}

For the district $\mathbf{D}^{\prime}=\left\{ X_{3}\right\} $, the
fixing set $\mathbf{W}=\left\{ X_{1},X_{2},X_{4}\right\} $, leading
to,

\begin{equation}
\begin{aligned}\Sigma^{2} & =\left(\mathrm{Simple}\circ\mathrm{Fix}_{X_{1}}\circ\mathrm{Fix}_{X_{2}}\circ\mathrm{Fix}_{X_{4}}\right)\left(\Sigma^{\mathcal{F}}\right)\\
 & =\left(\left\{ X_{1},X_{2},X_{3}\right\} ,\left\{ x_{3}\right\} ,\left\{ x_{3}:X_{1}X_{2}\to X_{3}\right\} \right)\\
 & =\mathrm{Ext}\left(\Sigma^{2}\right)\:\mathrm{Module}\left(X_{3}\right)=x_{3}.
\end{aligned}
\end{equation}

Finally, for the district $\mathbf{D}^{\prime}=\left\{ X_{4}\right\} $,
the fixing set $\mathbf{W}=\left\{ X_{1},X_{2},X_{3}\right\} $ leads
to,

\begin{equation}
\begin{aligned}\Sigma^{3} & =\left(\mathrm{Simple}\circ\mathrm{Hide}_{X_{2}}\circ\mathrm{Fix}_{X_{3}}\circ\mathrm{Fix}_{X_{1}}\right)\left(\Sigma^{\mathcal{F}}\right)\\
 & =\left(\left\{ X_{1},X_{2},X_{3},X_{4}\right\} ,\left\{ x_{4},x_{2}\right\} ,\left\{ x_{1}:X_{1}\to X_{1}^{2},x_{2}:X_{1}\to X_{2},x_{4}:X_{1}X_{2}X_{3}\to X_{4}\right\} \right),
\end{aligned}
\end{equation}
with exterior signature
\begin{equation}
\begin{aligned}\mathrm{Ext}\left(\Sigma^{3}\right) & =\left(\left\{ X_{1},X_{3},X_{4}\right\} ,\left\{ q\right\} ,\left\{ q:X_{1}X_{3}\to X_{4}\right\} \right),\,\mathrm{Module}\left(X_{4}\right)=q\\
q & =x_{4}\cdot\left(x_{2}\otimes\mathit{id}_{X_{1}}\otimes\mathit{id}_{X_{3}}\right)\cdot\left(x_{1}\otimes\mathit{id}_{X_{3}}\right),
\end{aligned}
\end{equation}
where the maximal exterior morphism $q$ has type,
\begin{equation}
\mathrm{dom}\left(q\right)=X_{1}X_{3}\to X_{1}X_{1}X_{3}\to X_{2}\to X_{1}X_{3}\to X_{4}=\mathrm{cod}\left(q\right).
\end{equation}

Combining the exterior signatures obtains

\begin{equation}
\begin{aligned}\Sigma_{X_{1}X_{3}X_{4}|\mathrm{do}\left(X_{2}\right)}^{\mathcal{G}} & =\mathrm{Ext}\left(\Sigma^{1}\right)\cup\mathrm{Ext}\left(\Sigma^{2}\right)\cup\mathrm{Ext}\left(\Sigma^{3}\right)\\
 & =\left(\left\{ X_{1},X_{2},X_{3},X_{4}\right\} ,\left\{ x_{1},x_{3},q\right\} ,\left\{ x_{1}:1\to X_{1}^{3},x_{3}:X_{1}X_{2}\to X_{3}^{2},q:X_{1}X_{3}\to X_{4}\right\} \right),
\end{aligned}
\end{equation}
from which we can compute the desired, syntactic interventional signature
by marginalization,

\begin{equation}
\begin{aligned}\Sigma_{X_{4}|\mathrm{do}\left(X_{2}\right)}^{\mathcal{G}} & =\mathrm{Hide}_{\left\{ X_{1},X_{3}\right\} }\left(\Sigma_{X_{1}X_{3}X_{4}|\mathrm{do}\left(X_{2}\right)}^{\mathcal{G}}\right)\\
 & =\left(\left\{ X_{1},X_{2},X_{3},X_{4}\right\} ,\left\{ x_{1},x_{3},q\right\} ,\left\{ x_{1}:1\to X_{1}^{2},x_{3}:X_{1}X_{2}\to X_{3},q:X_{1}X_{3}\to X_{4}\right\} \right).
\end{aligned}
\end{equation}

\section{Discussion}

In this paper, we have shown that purely syntactic causal identification
can be performed using relatively simple steps. We observe that the
simplicity of this approach largely arises from the process-centric
formulation of directed causal modelling and the fact that manipulations
of this model this can be expressed in terms of functions of the signature
of the category in which this model is represented. These steps are
unambiguous and therefore easily implemented in software.

Although our approach relies on chain factorization of the observed
process, we note that this is more of a mathematical convenience than
a restriction. An alternative development of our approach can use
\emph{comb disintegration} in place of the fixing operator described
here. This would lead to different forms of the resulting interventional
signatures which are, nonetheless equivalent exterior processes. Furthermore,
this approach could be extended to edge interventions, and we believe
this would be quite simple to implement. 

Finally, application to other, more elaborate forms of causal identification
such as conditional causal effects, those arising through edge interventions,
and more general forms of causal identification by combining multiple
causal models, would be valuable. It would be interesting to see the
extent to which the signature-based approach also simplifies the formulation
of existing algorithms for these problems.

\bibliographystyle{plainnat}
\bibliography{sig_id_3}

\end{document}